\documentclass[runningheads]{llncs}
\usepackage{graphicx}
\usepackage{todonotes}
\usepackage{hyperref}
\usepackage{cite}
\usepackage{bbm}
\usepackage{amsmath}
\usepackage{amssymb}
\usepackage{multirow}
\usepackage{algorithm,algorithmic}
\usepackage{comment}
\usepackage{lipsum}

\begin{document}
\title{Have you forgotten? A method to assess if machine learning models have forgotten data}
\titlerunning{A method to assess if machine learning models have forgotten data}
\author{Xiao Liu\inst{1} \and
Sotirios A. Tsaftaris\inst{1,2}} 
%index{Liu, Xiao}
%index{Tsaftaris, Sotirios}
\authorrunning{X. Liu et al.}

\institute{School of Engineering, University of Edinburgh, Edinburgh EH9 3FB, UK \and
The Alan Turing Institute, London, UK \\
\email{Xiao.Liu@ed.ac.uk, S.Tsaftaris@ed.ac.uk}}

\maketitle             
\setcounter{footnote}{0}
\begin{abstract}
In the era of deep learning, aggregation of data from several sources is a common approach to ensuring data diversity. Let us consider a scenario where several providers contribute data to a consortium for the joint development of a classification model (hereafter the target model), but, now one of the providers decides to leave.  This provider requests that their data (hereafter the query dataset) be removed from the databases but also that the model `forgets' their data. In this paper, for the first time, we want to address the challenging question of whether data have been forgotten by a model. We assume knowledge of the query dataset and the distribution of a model's output. We establish statistical methods that compare the target's outputs with outputs of models trained with different datasets. We evaluate our approach on several benchmark datasets (MNIST, CIFAR-10 and SVHN) and on a cardiac pathology diagnosis task using data from the Automated Cardiac Diagnosis Challenge (ACDC).  We hope to encourage studies on what information a model retains and inspire extensions in more complex settings.
\keywords{Privacy \and Statistical measure \and Kolmogorov-Smirnov}
\end{abstract}

\section{Introduction}
Deep learning requires lots of, diverse, data and in healthcare likely we will need to enlist several data providers (e.g. hospital authorities, trusts, insurance providers who own hospitals, etc) to ensure such data diversity. To develop models, data from several providers must thus be either centrally aggregated or leveraged within a decentralized federated learning scheme that does not require the central aggregation of data (e.g. \cite{barillot2006federating, ShellerREMB18,li2020federated,Li2019KCL,Roy2019BrainTorrentAP}).  Thus, several providers will contribute data for the development of a deep learning model e.g.\ to solve a simple classification task of normal vs. disease. Suddenly, one of the providers decides to leave and asks for the data to be deleted but more critically that \emph{the model `forgets' the data}.

\begin{figure}[t]
\includegraphics[width=\textwidth]{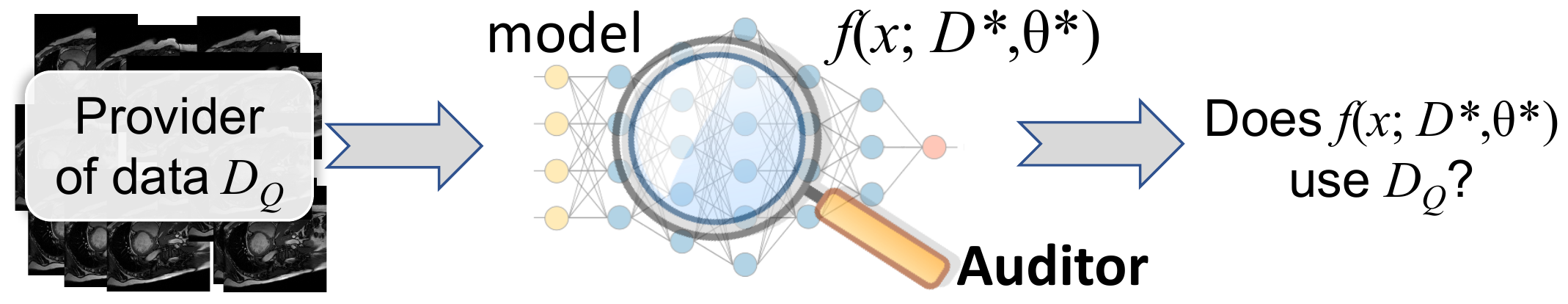}
\caption{Our goal. We aim to develop an approach that an `auditor' can use to ascertain whether a model $f(x;D^*;\theta)$ has used data $D_Q$ during training.} \label{fig:1:concept}
\end{figure}

Let us now assume that the model has \textit{indeed} not used the data and has `forgotten' them (we will not care how herein) but we want to \textit{verify} this. In other words, and as illustrated in Fig. \ref{fig:1:concept}, our problem is to assess whether the model has not used the data under question in the training set.

We consider that an \textit{auditor} will have access to the query dataset $D_Q$ and the model $f(\textbf{x};D^*,\theta^*)$, (trained on data $D^*$) to render a decision whether $f()$ retains information about $D_Q$. We assume that the training dataset $D^{*}$ is \textit{unknown} but both $D^*$ and $D_Q$ are sampled from a \textit{known} domain $\mathbbm{D}$ for a given \emph{known task}.\footnote{Knowledge of the task (e.g.\ detect presence of a pathology in cardiac MRI images), implies knowledge of the domain $\mathbbm{D}$ (e.g.\ the space of cardiac MRI images). Without this assumption, $D^*$ can be anything, rendering the problem intractable.} 
  We emphasize that we are \emph{not} proposing a method that forgets data, as for example the one in \cite{golatkar2019eternal}. We believe that assessing whether data have been forgotten is an important task that should come first (because without means to verify claims, anyone can claim that data have been forgotten).

Related work is limited. In fact, our communities (machine learning or medical image analysis) have not considered this problem yet in this form. The closest problem in machine learning is the \textit{Membership Inference Attack} (MIA) \cite{Shokri17, Carlini2019, Cherubin2019, Sablayrolles2018, Pyrgelis2019}. While these works borrow concepts from MIA, as we detail in the related work section, MIA sets different hypotheses that have considerable implications in returning false positives when data sources overlap. Our major novelty is setting appropriate test hypotheses that calibrate for data overlap.

Our approach builds on several key ideas. First, inspired by \cite{Sablayrolles2018}, we adopt the Kolmogorov-Smirnov (K-S) distance to compare statistically the similarity between the distribution of a model $f()$'s outputs and several purposely constructed reference distributions. To allow our approach to operate in a black-box setting (without precise knowledge of $f()$, we construct `shadow' models (inspired by \cite{Shokri17}) to approximate the target model $f()$; however, we train the shadow models on $D_Q$ but also on another dataset $D_C$ sampled from domain $\mathbbm{D}$ which does not overlap (in element-wise sense) $D_Q$. 

\noindent \textbf{Contributions}:
\begin{enumerate}
    \item To introduce a new problem in data privacy and retention to our community.
    \item To offer a solution that can be used to detect whether a model has forgotten specific data including the challenging aspect when data sources may overlap.
    \item Experiments in known image classification benchmark datasets to verify the effectiveness of our approach. Experiments on the pathology classification component of the ACDC simulating a healthcare-inspired scenario.
\end{enumerate}

\begin{comment}
\begin{figure}[t]
  \begin{minipage}[c]{0.4\textwidth}
\includegraphics[width=\textwidth]{figures/fig2-MIA_problem.pdf}
  \end{minipage}\hfill
  \begin{minipage}[c]{0.6\textwidth}
\caption{The problem of information overlap on a model $f()$ trained on $D^*$. If both  $D_Q$ and $D^*$ are drawn from domain $\mathbbm{D}$ they may  
overlap. An MIA algorithm on the point $x$ (not in the overlap region) will return, correctly, a true negative; however, on point $\textbf{x}'$ (in the region) will return a false positive: we cannot tell whether $x^*$ comes from $D_Q$ or $D^*$.  
} \label{fig:2:overlap_problem}
\end{minipage}
\end{figure}
\end{comment}

\section{Related work}
The purpose of \emph{Membership Inference Attack} is to learn which data were used to train a model. The hypothesis is that a model \emph{has} used some data and the goal is to find which part of the query data are in the training set.  
Here we briefly review key inspirational approaches.%
\footnote{We do not cover here the different task of making models and data more private, by means e.g.\ of \textit{differential privacy}. We point readers to surveys such as \cite{Gong2020, ji2014differential} and a recent (but not the only) example application in healthcare \cite{Li2019KCL}.}

In \cite{Shokri17} they train a model $f_{attack}$ that infers whether some data $\textbf{D}$ were used by $f$. $f_{attack}$ accepts as inputs the decisions of $f$ (the softmax outputs), $\textbf{D}$ and the ground-truth class. Their premise is that machine learning models often behave ``\textit{differently on the data that they were trained on vs. data that they `see' for the first time}''. Thus, the task reduces to training this attack model. They rely on being able to generate data that resemble $\textbf{D}$ (or not) and train several, different, `shadow' models that mimic $f$ to obtain outputs that can be used to train $f_{attack}$. We rely here on the same premise and use the idea of training shadow models of $f()$ to enable gray-box inference on $f()$ using only its outputs.

In \cite{Sablayrolles2018}, in the context of keeping machine learning competitions fair, using the same premise as \cite{Shokri17}, they propose a statistical approach to compare model outputs, and eschew the need for $f_{attack}$. They use Kolmogorov-Smirnov (K-S) distance to measure the statistical similarity between emissions of the classifier layer of a network between query data and a reference dataset to see whether models have used validation data to train with. However, their approach assumes that both query (validation set) and reference datasets, which in their context is the testing set, are known \textit{a priori}. We adopt (see Section \ref{sec:method}) the K-S distance as a measure of distribution similarity but different from \cite{Sablayrolles2018} we construct reference distributions to also calibrate for information overlap.

\noindent \textbf{Why our problem is not the same as MIA}:  
An MIA algorithm, by design, assumes that query data \emph{may} have been used to train a model. Whereas, we care to ascertain if a model has \emph{not} used the data.  These are different hypotheses which appear complementary but due to data source overlap have considerable implications on defining false positives and true negatives.\footnote{In fact, overlap will frequently occur in the real world. For example, datasets collected by different vendors can overlap if they collaborate with a same hospital or if a patient has visited several hospitals. Our method has been designed to address this challenging aspect of overlap.} When there is no information overlap between datasets, a MIA algorithm will return the right decision/answer. However, when $D_Q$ and $D^*$ statistically overlap (i.e.\ their manifolds overlap), for a sample both in $D_Q$ or $D^*$, a MIA algorithm will return a false positive even when the model was trained on $D^*$ alone. In other words, MIA \textit{cannot} tell apart data overlap. Since overlap between $D_Q$ and $D^*$ is likely and \textit{a priori} the auditor does not know it, the problem we aim to address has more \textit{stringent} requirements than MIA.  To address these requirements we assume that we can sample another dataset $D_C$ from $\mathbbm{D}$ (that we can control not to at least sample-wise overlap with $D_Q$) to train shadow models on $D_C$. 

\begin{comment}
\subsection{Measuring Data Overlap}
Measures to estimate overlap between distributions in high dimensions is not trivial.
Gaizer et. al.~\cite{Glazer2012LearningHR} peruses one-class support vector machines to find a 1-dimensional projection which can be used to estimate the Kolmogorov-Smirnov (KS) two-sample statistic.  Recently, MINE, a mutual information estimator using neural networks has been proposed \cite{pmlr-v80-belghazi18a}.
To use these metrics one needs \emph{both} $D_Q$ and $D^*$. However, if $D^*$ is known, inferring if $D_Q$ was used to train $f()$ becomes a trivial problem since we can, element-wise, compare $D_Q$ and $D^*$ directly without caring about $f()$. On the other extreme, strictly no information on $D^*$ makes the problem ill-posed. Thus, we consider a 'gray-box' setting with \textit{known $D_Q$, but $D^*$ unknown but sampled from domain $\mathbbm{D}$, which captures knowledge of the task of model $f()$}.
\end{comment}

\begin{algorithm}
\caption{The proposed method.}
\begin{algorithmic}
\label{AlgorithmMethod}
\renewcommand{\algorithmicrequire}{\textbf{Input:}}
\renewcommand{\algorithmicensure}{\textbf{Output:}}
\REQUIRE the target model $f(\textbf{x};D^*,\theta^*)$, query dataset $D_Q$ and dataset domain $\mathbbm{D}$.
\ENSURE answer to hypothesis if $f(\textbf{x};D^*,\theta^*)$ is trained with $D_Q$.
\STATE \textbf{Step 1.} Train $\widetilde{f}(\textbf{x};D_Q,\theta_Q)$ with the same design of $f(\textbf{x};D^*,\theta^*)$.
\STATE \textbf{Step 2a.} Sample the calibration dataset $D_C$ from domain $\mathbbm{D}$ but without overlapping (in sample-wise sense) $D_Q$.
\STATE \textbf{Step 2b.} Train $\widetilde{f}(\textbf{x};D_C,\theta_C)$ with the same design of $f(\textbf{x};D^*,\theta^*)$.
\STATE \textbf{Step 3.} Find $r(T_{D_Q|D_Q})$, $r(T_{D_Q|D^*})$ and $r(T_{D_Q|D_C})$ (Eq. \ref{EDFX}).
\STATE \textbf{Step 4.} Find $KS(r(T_{D_Q|D_Q}), r(T_{D_Q|D^*}))$, $KS(r(T_{D_Q|D_Q}), r(T_{D_Q|D_C}))$ (Eqs. \ref{OneSampleK-SDistance}, \ref{OneSampleK-SDistance2}).
\STATE \textbf{Step 5.} Find $\rho = \frac{KS(r(T_{D_Q|D_Q}), r(T_{D_Q|D^*}))}{ KS(r(T_{D_Q|D_Q}), r(T_{D_Q|D_C}))}$ (Eq. \ref{eq:ratio}).
\RETURN If $\rho \geq 1$, the target model has forgotten $D_Q$. Otherwise ($\rho < 1$), has not.
%\RETURN answer according to $\rho$.
\end{algorithmic} 
\end{algorithm}

\section{Proposed method}
\label{sec:method}
We measure similarity of distributions to infer whether a model is trained with the query dataset $D_Q$, i.e. if $D^*$ has $D_Q$ or not. Note that this is not trivial since $D^*$ is \textit{unknown}. In addition, the possible overlap between $D^*$ and $D_Q$ introduces more challenges as we outlined above. To address both, we introduce another dataset $D_C$ sampled from domain $\mathbbm{D}$ but without overlapping (in element-wise sense) $D_Q$. The method is summarized in Algorithm \ref{AlgorithmMethod} with steps detailed below.

\noindent \textbf{Notation:} We will consider $\textbf{x}$, a tensor, the input to the model e.g. an image. We will denote the \emph{target} model as $f(\textbf{x}; D^*,\theta^*)$ that is trained on dataset $D^*$ parametrised by $\theta^*$. Similarly, we define \emph{query} $\widetilde{f}(\textbf{x}; D_Q,\theta_Q)$ and \emph{calibration} models $\widetilde{f}(\textbf{x}; D_C,\theta_C)$, where both models share model design with the target model. If $D_Q$ has $N$ samples ($N \gg 1$), we use $y^{\{i\}}, i=\{1,\cdots,N\}$, a scalar, to denote the ground-truth label of data $\textbf{x}^{\{i\}}$ of $D_Q$. We denote with $\textbf{t}^{\{i\}}, i=\{1,\cdots,N\}$, a $M$-dimensional vector, the \emph{output} of a model with input $\textbf{x}^{\{i\}}$. Hence, $\textbf{t}^{\{i\}}[y^{\{i\}}]$ denotes the output of the model $\textbf{t}^{\{i\}}$ for the ground-truth class $y^{\{i\}}$ (confidence score). We further define the $N\times 1$ vector $\textbf{c}_{D_Q|D^*}$, which contains the values of $\textbf{t}^{\{i\}}[y^{\{i\}}]$, $i=\{1,\cdots,N\}$ in an increasing order.
We define as $T_{D_Q|D^*}$ the $N\times M$ matrix that contains all outputs of the target model $f(\textbf{x}; D^*,\theta^*)$ that is tested with $D_Q$. Similarly, we define $T_{D_Q|D_Q}$ and $T_{D_Q|D_C}$ for the outputs of $\widetilde{f}(\textbf{x}; D_Q,\theta_Q)$ and $\widetilde{f}(\textbf{x}; D_C,\theta_C)$, respectively,  when both are tested with $D_Q$.

\subsection{Assumptions on $f()$}
We follow a gray-box scenario: access to the output of the model $f(\textbf{x}; D^*,\theta^*)$ before any thresholds and knowledge of design (e.g.\ it is a neural network) but not of the parameters (e.g.\ weights) $\theta^*$. (Model providers typically provide the best (un-thresholded) value as surrogate of (un)certainty.) 

% \todo{Xiao I think we have an error on eq 1 (definition of cdf, and consequently confusion on eq 2. I dont think it should be the $n-th$ element but the c element no? In my view the cdf $F(t)$ is the cummulative sum over data $N$, how many data $ x \le t$.
% I think this is what  $\textbf{t}^{\{i\}}(y^{\{i\}}) \le c$ refer to? see \url{https://en.wikipedia.org/wiki/Kolmogorov\%E2\%80\%93Smirnov_test}}

\subsection{Kolmogorov-Smirnov distance between distributions}
We denote as $r(T_{D_Q|D^*})$ (a $N \times 1$ vector) the empirical cumulative distribution (cdf) of the output of the target model when tested with $D_Q$. The  $n^{th}$ element $r_n(T_{D_Q|D^*})$ is defined as:
\begin{equation}
    \label{EDFX}
    r_n(T_{D_Q|D^*}) = \frac{1}{N}\sum_{i=1}^{N}I_{[-\infty, \textbf{c}_{D_Q|D^*}^{\{n\}}]}(\textbf{c}^{\{i\}}_{D_Q|D^*}),
\end{equation}
where $I_{[-\infty,\textbf{c}_{D_Q|D^*}^{\{n\}}]}(\textbf{c}^{\{i\}}_{D_Q|D^*})$ the indicator function, is equal to 1 if $\textbf{c}^{\{i\}}_{D_Q|D^*}\leq \textbf{c}^{\{n\}}_{D_Q|D^*}$ and equal to 0 otherwise.
% \begin{equation}
%     \label{IndicatorFunction}
%     I(\textbf{t}^{\{i\}}[y^{\{i\}}]) = \Bigg\{
%     \begin{aligned}
%     &1, \quad \textbf{t}^{\{i\}}(y^{\{i\}}) \le c, \\&0, \quad \text{otherwise}.
%     \end{aligned}
% \end{equation}

%If there is an external reference dataset for $D_Q$, the method in \cite{Sablayrolles2018} can be leveraged. 
%In our problem setting, $D^*$ is unknown and the method in \cite{Sablayrolles2018} requia reference dataset. 
Our next step is to create a proper empirical distribution as  reference to compare against $r(T_{D_Q|D^*})$. Motivated by \cite{Shokri17}, we propose to train a query model $\widetilde{f}(\textbf{x};D_Q,\theta_Q)$ with the same model design as the target model. We then obtain (as previously) the output cumulative distribution $r(T_{D_Q|D_Q})$. We propose to use $r(T_{D_Q|D_Q})$ as the reference distribution of dataset $D_Q$. Hence, measuring the similarity between $r(T_{D_Q|D^*})$ with $r(T_{D_Q|D_Q})$ can inform on the relationship  between $D_Q$ and $D^*$, which is extensively explored in the field of dataset bias \cite{torralba2011unbiased}. Following \cite{Sablayrolles2018}, we use Kolmogorov–Smirnov (K-S) distance as a measure of the similarity between the two empirical distributions. 

K-S distance was first used in \cite{feller2015kolmogorov} to compare a sample with a specific reference distribution or to compare two samples. For our purpose we will peruse the two-sample K-S distance, which is given by: 
\begin{equation}
    \label{OneSampleK-SDistance}
    KS(r(T_{D_Q|D_Q}), r(T_{D_Q|D^*})) = sup|r(T_{D_Q|D_Q}) - r(T_{D_Q|D^*})|_1,
\end{equation}
where $sup$ denotes the largest value. K-S distance $\in [0,1]$  with lower values pointing to greater similarity between $r(T_{D_Q|D^*})$ and $r(T_{D_Q|D_Q})$.

If $D^*$ contains $D_Q$, the K-S distance between $r(T_{D_Q|D_Q})$ and $r(T_{D_Q|D^*})$ will be very small and $\approx 0$, i.e. $KS(r(T_{D_Q|D_Q}), r(T_{D_Q|D^*})) \approx 0$. On the contrary, if $D^*$ has no samples from $D_Q$, then the value of $KS(r(T_{D_Q|D_Q}), r(T_{D_Q|D^*}))$ depends on the statistical overlap of $D^*$ and $D_Q$. However, this overlap is challenging to measure because the training dataset $D^*$ is assumed unknown (and we cannot approximate statistical overlap with element-wise comparisons).

\subsection{Calibrating for data overlap}

We assume the training data $D^{*}$ of model $f(\textbf{x};D^*,\theta^*)$ are sampled from a domain $\mathbbm{D}$. To calibrate against overlap between $D^*$ and $D_Q$, we sample a \emph{calibration} dataset $D_C$ from $\mathbbm{D}$ but ensure that no samples in $D_Q$ are included in $D_C$ (by sample-wise comparisons) and train the calibration model $\widetilde{f}(\textbf{x}; D_C,\theta_C)$. Inference on model $\widetilde{f}(\textbf{x}; D_C,\theta_C)$ with $D_Q$, the output empirical cumulative distribution $r(T_{D_Q|D_C})$ can be calculated in a similar fashion as Eq. \ref{EDFX}. To compare the similarity between $D_Q$ and $D_C$, we calculate:
\begin{equation}
    \label{OneSampleK-SDistance2}
    KS(r(T_{D_Q|D_Q}), r(T_{D_Q|D_C})) = sup|r(T_{D_Q|D_Q}) - r(T_{D_Q|D_C})|_1.
\end{equation}

We argue that Eq.~\ref{OneSampleK-SDistance2} can be used to calibrate the overlap  and inform if $D^*$ contains $D_Q$.\footnote{As mentioned previously we cannot measure statistical overlap between $D^*$ and $D_C$ (or $D_Q$) since $D^*$ is unknown. Furthermore, statistical overlap in high-dimensional spaces is not trivial~\cite{Glazer2012LearningHR,pmlr-v80-belghazi18a}, and hence we want to avoid it.} We discuss this for two scenarios: \\
\textbf{1.  $D^*$ does not include any samples of $D_Q$.} Since $D_C$ and $D_Q$ do not overlap in a element-wise sense by construction, then data of $D^*$ should have higher probability of statistical overlap with $D_C$. In other words, the calibration model can be used to mimic the target model. By sampling more data from domain $\mathbbm{D}$ as $D_C$, there will be more statistical overlap between $D_C$ and $D_Q$. Hence, the statistical overlap between a large-size dataset $D_C$ and $D_Q$ can be used to approximate the (maximum) possible overlap between $D^*$ and $D_Q$ i.e. $KS(r(T_{D_Q|D_Q}), r(T_{D_Q|D^*})) \geq KS(r(T_{D_Q|D_Q}), r(T_{D_Q|D_C}))$.

\noindent \textbf{2. $D^*$ contains samples of $D_Q$.} In this case, samples in $D_Q$ that are also in $D^*$ will introduce statistical overlap between $D_Q$ and $D^*$, which is well demonstrated in \cite{Sablayrolles2018}. Based on our experiments, we find that the overlap between $D_Q$ and $D^*$ is consistently less than that of $D_Q$ and $D_C$ i.e. $KS(r(T_{D_Q|D_Q}), r(T_{D_Q|D^*})) < KS(r(T_{D_Q|D_Q}), r(T_{D_Q|D_C}))$.

Thus, we propose to use $KS(r(T_{D_Q|D_Q}), r(T_{D_Q|D_C}))$ as a data-driven indicator for detecting if $D^*$ contains $D_Q$. To quantify how much the target model has forgotten about $D_Q$, we capture the two inequalities above in the ratio:
\begin{equation}
\label{eq:ratio}
    \rho = \frac{KS(r(T_{D_Q|D_Q}), r(T_{D_Q|D^*}))}{KS(r(T_{D_Q|D_Q}),r(T_{D_Q|D_C}))}.
\end{equation}
In the first scenario i.e. $D^*$ does not include any samples of $D_Q$, the inequality $KS(r(T_{D_Q|D_Q}), r(T_{D_Q|D^*})) \geq KS(r(T_{D_Q|D_Q}), r(T_{D_Q|D_C}))$ translates to $\rho \geq 1$. Instead, in the second scenario, $\rho < 1$. Thus, if $\rho \geq 1$, this implies the target model has forgotten $D_Q$. On the contrary, if $\rho < 1$, implies the target model has not forgotten $D_Q$.

\section{Experiments}
\label{sec:Experiments}
We first perform experiments on benchmark datasets such as MNIST \cite{lecun1998gradient}, SVHN \cite{netzer2011reading} and CIFAR-10 \cite{krizhevsky2009learning} to verify the effectiveness of the proposed method. Then, we test our method on the ACDC dataset using the pathology detection component of the challenge \cite{bernard2018deep}. All classifiers in our experiments are well-trained (adequate training accuracy) such that the output statistics well approximate the input statistics. 
% All model designs are included as Supplementary Material.
\subsection{Benchmark datasets - MNIST, SVHN and CIFAR-10}
\label{sec:BenchmarkDataset}

\noindent\textbf{Benchmark datasets:}  MNIST contains 60,000 images of 10 digits with image size $28\times 28$. Similar to MNIST, SVHN has over 600,000 digit images obtained from house numbers in Google Street view images. The image size of SVHN is $32\times 32$. Since both datasets are for the task of digit recognition/classification we consider to belong to the same domain $\mathbbm{D}$. We use CIFAR-10 as an \textit{out of domain} $\mathbbm{D}$ dataset to validate the method. CIFAR-10 has 60,000 images (size $32\times 32$) of 10-class objects (airplane, bird, etc). To train our models, we preprocess images of all three datasets to gray-scale and rescale to size $28\times 28$. 

\noindent\textbf{Implementation details:} In terms of the model architecture, we utilize a three-block classifier with one convolution block, one linear block and an output linear layer.
The convolution block contains one convolution layer with 32 channels, kernel size 7, stride size 1 and padding size 3 and uses ReLU \cite{nair2010rectified} as the activation function and batch normalization \cite{ioffe2015batch} is also included. The last layer of this convolution block is max pooling \cite{krizhevsky2012imagenet} with stride size 2 and kernel size 2 to further downsample the input feature maps. The linear block has one fully-connected linear layer and a ReLU activation function. Finally, the output layer is also fully-connected layer that maps a 1,024-dimensional input vector to 10-dimensional output. We use cross entropy function as the objective of training. Adam optimizer \cite{kingma2014adam} with $\beta1=0.5$, $\beta2=0.999$ is used to train all the models with 15 epochs.

\begin{table}[t]
\centering
\caption{Benchmark datasets results. The query dataset $D_Q$ is MNIST. The training dataset $D^*$ has different variants as listed below. The K-S distances are calculated with Eq. \ref{OneSampleK-SDistance}. $D_C$ for defining $\rho$ is SVHN. Bold number in the table is the threshold.}\label{tab1}
\begin{tabular}{|c|c|c|c|c|c|c|c|c|}
\hline
\multirow{ 2}{*}{Training dataset ($D^*$)} & \multirow{ 2}{*}{MNIST} & \multicolumn{3}{c|}{\%SVHN} & \multicolumn{3}{c|}{SVHN + \%MNIST} & \multirow{ 2}{*}{CIFAR-10} \\ \cline{3-8}
 & & 50\% & 75\% & 100\% & 10\% & 50\% & 100\% &\\
\hline
$KS(r(t_{D_Q|D_Q}), r(t_{D_Q|D^*}))$ & 0.049 & 0.669 & 0.652 & \textbf{0.596} & 0.056 & 0.039 & 0.029 & 0.957 \\
\hline
$\rho$      & 0.082  & 1.122 & 1.094 & 1.000 & 0.094 & 0.065 & 0.049 & 1.606\\
\hline
\end{tabular}
\end{table}

\noindent\textbf{Results and discussion:} We consider MNIST as $D_Q$. Experimental results are listed in Table \ref{tab1}. We first verify if our method can detect the simple case where the target model has been fully trained with $D_Q$. According to our findings (first column of Table \ref{tab1}) as expected $\rho=0.082 \ll 1$. Other cases where we assume $D^*$ is drawn only from $D_C$ (SVHN) give $\rho > 1$. In the challenging setting when the target model was trained with a mix of data from SVHN and parts of MNIST, $\rho \sim 0$.  We draw attention to the case of SVHN+$\%100$MNIST. $\rho=0.049$ indicates that the target definitely has not forgotten the query dataset.

For exploration purposes, we consider the extreme case of $D^*$ not in the domain of $D_Q$. We train a target model with CIFAR-10 which has little statistical information overlap with MNIST. Hence, we would expect our method to return a $\rho > 1$, which agrees with the result $\rho = 1.606$ (highest in the table).

Next we assess the scenarios that $D^*$ does not include any samples of $D_Q$. We consider SVHN as $D_C$ and train target models with $D^*$, subsets of SVHN. The results $\rho$ of $50\%$ and $75\%$ SVHN are 1.122 and 1.094 that both higher than $\rho = 1$. Higher $\rho$ points to lower information overlap between $D_Q$ and $D^*$. Hence, the large-size $D_C$ i.e. $\%100$SVHN has the largest statistical overlap with $D_Q$.
%Hence, these results stand by our argument.

To check the sensitivity of our method when $D^*$ contains part of $D_Q$, we train two models with SVHN and part of MNIST ($10\%$ and $50\%$). We observe that both have $\rho < 1$. Specifically, including only $10\%$ of MNIST causes $(1-0.094)\div 1$ = $90.6\%$ drop of $\rho$ value. This suggests that if the training dataset contains part of $D_Q$, our method can still accurately detect these cases.

Some uncertainty may arise if $\rho$ is hovering around 1. However, we did not observe such cases when $D^*$ contains samples from $D_Q$ in our extensive experiments. We advise to run multiple runs of experiments to statistically eliminate such uncertainty if needed.

\begin{figure}[t]
\centering
\includegraphics[width=0.9\textwidth]{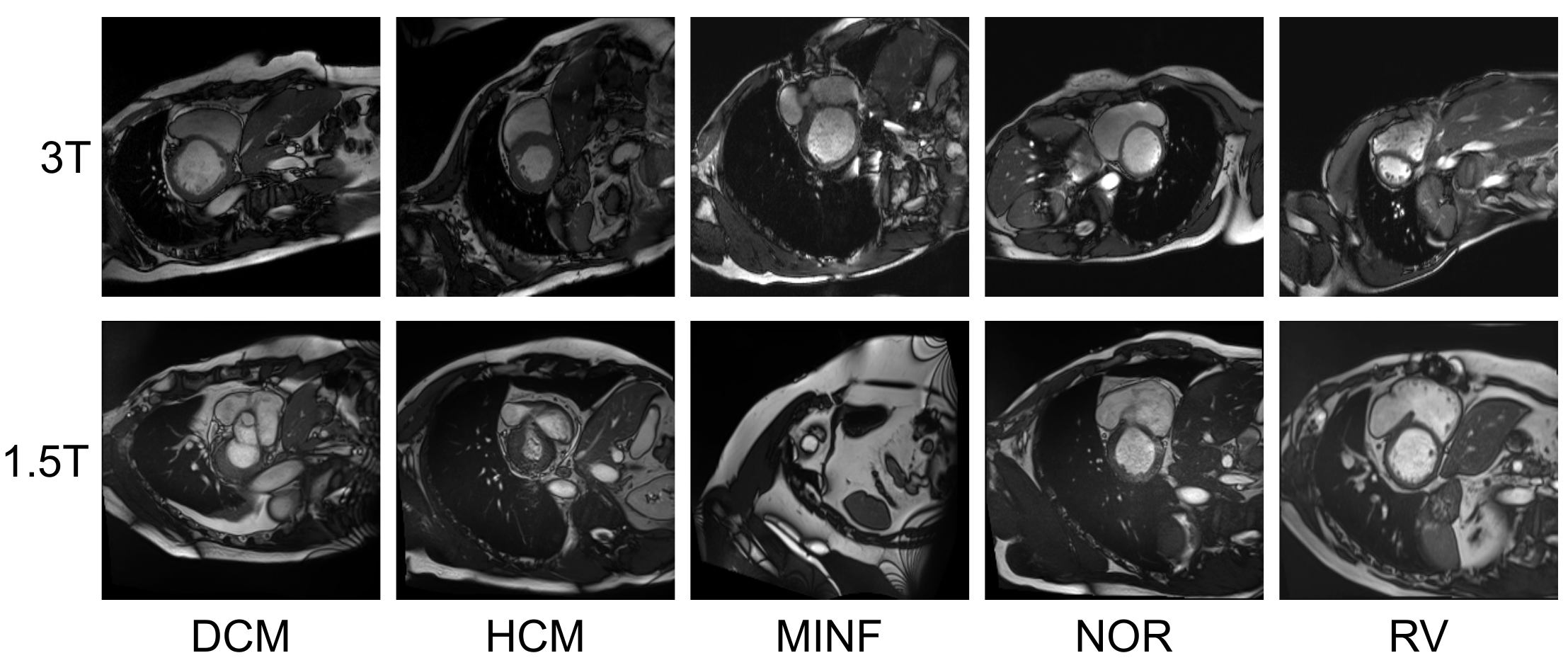}
\caption{Example images of ACDC 1.5T and ACDC 3T datasets. DCM: dilated cardiomyopathy. HCM: hypertrophic cardiomyopathy. MINF: myocardial infarction. NOR: normal subjects. RV: abnormal right ventricle.} \label{fig:3:ACDC}
\end{figure}

\subsection{Medical dataset - ACDC}
\noindent\textbf{ACDC dataset:} Automated Cardiac Diagnosis Challenge (ACDC) dataset is composed of cardiac images of 150 different patients that are acquired with two MRI scanners of different magnetic strengths, i.e. 1.5T and 3.0T. For each patient, ACDC provides 4-D data (weight, height, diastolic phase instants, systolic phase instants). To simulate several data providers we split the original training data of ACDC into two datasets: we consider the images at 1.5T field strength as one source and those at 3T as the other. Overall, ACDC 1.5T has 17,408 images and ACDC 3T has 7,936 images (Examples are shown in Fig. \ref{fig:3:ACDC}.). For both datasets, four classes of pathology are provided and one for normal subjects (for a total of 5 classes). We randomly crop all images to $128\times 128$ during training. 

\noindent\textbf{Implementation details:} All classifiers have same model design. Specifically, there are 4 convolution blocks followed with 1 linear output layer. The first convolution block contains a convolution layer with 32 channels, kernel size 7, stride size 1 and padding size 3. The other three convolution blocks have 64, 128, 128 channels with kernel size 4, stride size 2 and padding size 1. Cross entropy function is used as the objective of training. Adam optimizer \cite{kingma2014adam} with $\beta1=0.5$, $\beta2=0.999$ is used to train all the models with 15 epochs.

\noindent\textbf{Results and discussion:} We consider ACDC 3T as $D_Q$ and assume the domain $\mathbbm{D}$ spans the space of cine images of the heart in 1.5T or 3T strength.  ACDC 1.5T is the calibration dataset $D_C$. Results are shown in Table~\ref{tab2}.

We first perform experiments to verify if our method can correctly detect whether $D_Q$ is in $D^*$. Similar to benchmark datasets, all models trained with ACDC 3T or part of ACDC 3T have $\rho < 1$. Other models trained without any data from ACDC 3T have $\rho \geq 1$. Hence, for ACDC dataset, our method returns correct answers in all experiments.

According to $\%$ACDC 1.5T results, a large-size dataset $D_C$ ($\%100$ ACDC 1.5T) achieves lower K-S distance compared to $50\%$ and $75\%$ ACDC 1.5T. This supports our discussion of the scenario when $D^*$ does not include any samples of $D_Q$. Note that when mixing all data of ACDC 1.5T and only $10\%$ of 3T, the K-S distance drops $(1-0.750)\div 1 = 25\%$. It suggests even if part of $D_Q$ is in $D^*$, the proposed method is still sensitive to detect such case.

\begin{table}[t]
\centering
\caption{ACDC dataset results. The query dataset $D_Q$ is ACDC 3T. The training dataset $D^*$ has different variants as listed below. The K-S distances are calculated with Eq. \ref{OneSampleK-SDistance}. $D_C$ for defining $\rho$ is ACDC 1.5T. Bold number in the table is the threshold.}\label{tab2}
\begin{tabular}{|c|c|c|c|c|c|c|c|}
\hline
\multirow{ 2}{*}{Training dataset ($D^*$)} & \multirow{ 2}{*}{ACDC 3T} & \multicolumn{3}{c|}{\%ACDC 1.5T} & \multicolumn{3}{c|}{ACDC 1.5T + \% 3T} \\ \cline{3-8}
 & & 50\% & 75\% & 100\% & 10\% & 50\% & 100\% \\
\hline
$KS(r(T_{D_Q|D_Q}), r(T_{D_Q|D^*}))$ & 0.036 & 0.885 & 0.793 & \textbf{0.772} & 0.579 & 0.529 & 0.103 \\
\hline
$\rho$       & 0.047 & 1.146 & 1.027 & 1.000 & 0.750 & 0.685 & 0.133 \\
\hline
\end{tabular}
\end{table}

\section{Conclusion}
We introduced an approach that uses Kolmogorov-Smirnov (K-S) distance to detect if a model has used/forgotten a query dataset. Using the K-S distance we can obtain statistics about the output distribution of a target model without knowing the weights of the model. Since the model's training data are unknown we train new (shadow) models with the query dataset and another calibration dataset. By comparing the K-S values we can ascertain if the training data contain data from the query dataset even for the difficult case where data sources can overlap. We showed experiments in classical classification benchmarks but also classification problems in medical image analysis.
We did not explore the effect of the query dataset's sample size but the cumulative distribution remains a robust measure even in small sample sizes. Finally, extensions to segmentation or regression tasks and further assessment whether differential-privacy techniques help protect data remain as future work. 

\section{Acknowledgment}
This work was supported by the University of Edinburgh by a PhD studentship. 
This work was partially supported by the Alan Turing Institute under the EPSRC grant EP/N510129/1.  S.A. Tsaftaris acknowledges the support of the Royal Academy of Engineering and the Research Chairs and Senior Research Fellowships scheme and the [in part] support of the Industrial Centre for AI Research in digital Diagnostics (iCAIRD) which is funded by Innovate UK on behalf of UK Research and Innovation (UKRI) [project number: 104690] (\url{https://icaird.com/}).

\bibliographystyle{splncs04}
\bibliography{references}

\begin{thebibliography}{10}
\providecommand{\url}[1]{\texttt{#1}}
\providecommand{\urlprefix}{URL }
\providecommand{\doi}[1]{https://doi.org/#1}

\bibitem{barillot2006federating}
Barillot, C., Benali, H., Dojat, M., Gaignard, A., Gibaud, B.,
  Kinkingn{\'e}hun, S., Matsumoto, J., P{\'e}l{\'e}grini-Issac, M., Simon, E.,
  Temal, L.: Federating distributed and heterogeneous information sources in
  neuroimaging: the neurobase project. Studies in health technology and
  informatics  \textbf{120}, ~3 (2006)

\bibitem{pmlr-v80-belghazi18a}
Belghazi, M.I., Baratin, A., Rajeshwar, S., Ozair, S., Bengio, Y., Courville,
  A., Hjelm, D.: Mutual information neural estimation. In: Dy, J., Krause, A.
  (eds.) Proceedings of the 35th International Conference on Machine Learning.
  Proceedings of Machine Learning Research, vol.~80, pp. 531--540. PMLR,
  Stockholmsmässan, Stockholm Sweden (10--15 Jul 2018)

\bibitem{bernard2018deep}
Bernard, O., Lalande, A., Zotti, C., Cervenansky, F., Yang, X., Heng, P.A.,
  Cetin, I., Lekadir, K., Camara, O., Ballester, M.A.G., et~al.: Deep learning
  techniques for automatic mri cardiac multi-structures segmentation and
  diagnosis: is the problem solved? IEEE transactions on medical imaging
  \textbf{37}(11),  2514--2525 (2018)

\bibitem{Carlini2019}
Carlini, N., Liu, C., Erlingsson, U., Kos, J., Song, D.: The secret sharer:
  Evaluating and testing unintended memorization in neural networks. In:
  Proceedings of the 28th USENIX Conference on Security Symposium. pp.
  267--284. SEC'19, USENIX Association, Berkeley, CA, USA (2019)

\bibitem{Cherubin2019}
Cherubin, G., Chatzikokolakis, K., Palamidessi, C.: {F-BLEAU: Fast Black-box
  Leakage Estimation}  (feb 2019), \url{http://arxiv.org/abs/1902.01350}

\bibitem{feller2015kolmogorov}
Feller, W.: On the {Kolmogorov--Smirnov} limit theorems for empirical
  distributions. In: Selected Papers I, pp. 735--749. Springer (2015)

\bibitem{Glazer2012LearningHR}
Glazer, A., Lindenbaum, M., Markovitch, S.: Learning high-density regions for a
  generalized kolmogorov-smirnov test in high-dimensional data. In: NIPS (2012)

\bibitem{golatkar2019eternal}
Golatkar, A., Achille, A., Soatto, S.: Eternal sunshine of the spotless net:
  Selective forgetting in deep networks (2019)

\bibitem{Gong2020}
{Gong}, M., {Xie}, Y., {Pan}, K., {Feng}, K., {Qin}, A.K.: A survey on
  differentially private machine learning [review article]. IEEE Computational
  Intelligence Magazine  \textbf{15}(2),  49--64 (2020)

\bibitem{ioffe2015batch}
Ioffe, S., Szegedy, C.: Batch normalization: Accelerating deep network training
  by reducing internal covariate shift. In: International Conference on Machine
  Learning. pp. 448--456 (2015)

\bibitem{ji2014differential}
Ji, Z., Lipton, Z.C., Elkan, C.: Differential privacy and machine learning: a
  survey and review (2014)

\bibitem{kingma2014adam}
Kingma, D.P., Ba, J.L.: Adam: A method for stochastic gradient descent. In:
  ICLR: International Conference on Learning Representations (2015)

\bibitem{krizhevsky2009learning}
Krizhevsky, A., Hinton, G., et~al.: Learning multiple layers of features from
  tiny images  (2009)

\bibitem{krizhevsky2012imagenet}
Krizhevsky, A., Sutskever, I., Hinton, G.E.: Imagenet classification with deep
  convolutional neural networks. In: Advances in neural information processing
  systems. pp. 1097--1105 (2012)

\bibitem{lecun1998gradient}
LeCun, Y., Bottou, L., Bengio, Y., Haffner, P.: Gradient-based learning applied
  to document recognition. Proceedings of the IEEE  \textbf{86}(11),
  2278--2324 (1998)

\bibitem{li2020federated}
Li, T., Sahu, A.K., Talwalkar, A., Smith, V.: Federated learning: Challenges,
  methods, and future directions. IEEE Signal Processing Magazine
  \textbf{37}(3),  50--60 (2020)

\bibitem{Li2019KCL}
Li, W., Milletar{\`i}, F., Xu, D., Rieke, N., Hancox, J., Zhu, W., Baust, M.,
  Cheng, Y., Ourselin, S., Cardoso, M.J., Feng, A.: Privacy-preserving
  federated brain tumour segmentation. In: Suk, H.I., Liu, M., Yan, P., Lian,
  C. (eds.) Machine Learning in Medical Imaging. pp. 133--141. Springer
  International Publishing, Cham (2019)

\bibitem{nair2010rectified}
Nair, V., Hinton, G.E.: Rectified linear units improve restricted boltzmann
  machines. In: Proceedings of the 27th international conference on machine
  learning (ICML-10). pp. 807--814 (2010)

\bibitem{netzer2011reading}
Netzer, Y., Wang, T., Coates, A., Bissacco, A., Wu, B., Ng, A.Y.: Reading
  digits in natural images with unsupervised feature learning  (2011)

\bibitem{Pyrgelis2019}
Pyrgelis, A., Troncoso, C., {De Cristofaro}, E.: {Under the Hood of Membership
  Inference Attacks on Aggregate Location Time-Series}  (feb 2019),
  \url{http://arxiv.org/abs/1902.07456}

\bibitem{Roy2019BrainTorrentAP}
Roy, A.G., Siddiqui, S., P{\"o}lsterl, S., Navab, N., Wachinger, C.:
  Braintorrent: A peer-to-peer environment for decentralized federated learning
   (may 2019), \url{http://arxiv.org/abs/1905.06731}

\bibitem{Sablayrolles2018}
Sablayrolles, A., Douze, M., Schmid, C., J{\'{e}}gou, H.:
  {D$\backslash$'ej$\backslash$`a Vu: an empirical evaluation of the
  memorization properties of ConvNets}  (sep 2018),
  \url{http://arxiv.org/abs/1809.06396}

\bibitem{ShellerREMB18}
Sheller, M.J., Reina, G.A., Edwards, B., Martin, J., Bakas, S.:
  Multi-institutional deep learning modeling without sharing patient data: {A}
  feasibility study on brain tumor segmentation. In: Brainlesion: Glioma,
  Multiple Sclerosis, Stroke and Traumatic Brain Injuries - 4th Int. Workshop,
  BrainLes 2018, Held in Conjunction with {MICCAI} 2018, Granada, Spain, Sept.
  16, 2018, Part {I}. pp. 92--104 (2018)

\bibitem{Shokri17}
{Shokri}, R., {Stronati}, M., {Song}, C., {Shmatikov}, V.: Membership inference
  attacks against machine learning models. In: 2017 IEEE Symposium on Security
  and Privacy (SP). pp. 3--18 (May 2017). \doi{10.1109/SP.2017.41}

\bibitem{torralba2011unbiased}
Torralba, A., Efros, A.: Unbiased look at dataset bias. In: Proceedings of the
  2011 IEEE Conference on Computer Vision and Pattern Recognition. pp.
  1521--1528 (2011)

\end{thebibliography}

\end{document}